\RequirePackage{fix-cm}
\documentclass[twocolumn]{svjour3}          % twocolumn
\smartqed  % flush right qed marks, e.g. at end of proof
\usepackage{graphicx}

\usepackage{url}
\graphicspath{{./figures/}}%declare the path where your graphic files are
\usepackage{subfigure} % --- package 'subfig' will make the caption of tables/figure in the same font with the context, replace it with 'subfigure' and using 'subfigure[]' instead of 'subfloat[]'
\usepackage{amsfonts}
\usepackage{amsmath}
\usepackage{algorithm, algorithmic}
\usepackage[numbers,sort&compress]{natbib} %---result in abnormal of the reference font--solution: added '\footnotesize' before the reference
\usepackage{tabularx}
\begin{document}

\title{Stitching Videos from a Fisheye Lens Camera and a Wide-Angle Lens Camera for Telepresence Robots%\thanks{Grants or other notes
}


\author{Yanmei Dong, Mingtao Pei, Lijia Zhang, Bin Xu, Yuwei Wu, \\and Yunde Jia}


\institute{The authors are with Beijing Laboratory of Intelligent Information Technology, School of Computer Science, Beijing Institute of Technology, Beijing, 100081, P.R. China \\
              \email{\{dongyanmei, peimt, zhanglijia, xubinak47, wuyuwei, jiayunde\}@bit.edu.cn}           %  \\
}

\date{Received: date / Accepted: date}

\maketitle

\begin{abstract}
Many telepresence robots are equipped with a forward-facing camera for video communication and a downward-facing camera for navigation. In this paper, we propose to stitch videos from the FF-camera with a wide-angle lens and the DF-camera with a fisheye lens for telepresence robots. We aim at providing more compact and efficient visual feedback for the user interface of telepresence robots with user-friendly interactive experiences. To this end, we present a multi-homography-based video stitching method which stitches videos from a wide-angle camera and a fisheye camera. The method consists of video image alignment, seam cutting, and image blending. We directly align the wide-angle video image and the fisheye video image based on the multi-homography alignment without calibration, distortion correction, and unwarping procedures. Thus, we can obtain a stitched video with shape preservation in the non-overlapping regions and alignment in the overlapping area for telepresence. To alleviate ghosting effects caused by moving objects and/or moving cameras during telepresence robot driving, an optimal seam is found for aligned video composition, and the optimal seam will be updated in subsequent frames, considering spatial and temporal coherence. The final stitched video is created by image blending based on the optimal seam. We conducted a user study to demonstrate the effectiveness of our method and the superiority of telepresence robots with a stitched video as visual feedback.
\end{abstract}

\section{Introduction}
\label{sec: introduction}
A telepresence robot is a form of a video conferencing device mounted on a mobile robot, which allows a remote operator to teleoperate the robot as his/her embodiment to actively telecommunicate with local persons \cite{jia2015telepresence}. In recent years, telepresence robots are increasingly common in various everyday contexts, such as office environments \cite{neustaedter2016to}, %fairchild2017mixed,
remote education \cite{Meyer2016Mobile}, technical mediation \cite{bae2018public}, %Rudolph2017Integrating,
elderly people support \cite{Cesta2016Long}, and residential care \cite{niemela2019towards}. %health care \cite{marini2015effect}.
Many existing telepresence robots \cite{kristoffersson2013review, Beam2017, Double2017} are equipped with a forward-facing camera (FF-camera) for video communication and a downward-facing camera (DF-camera) for navigation. The two cameras provide two live videos displayed on two corresponding windows in the GUI as visual feedback. However, in testing visual feedback with these two video windows, we found that two video windows could introduce some confusion over a local environment \cite{jia2015telepresence}. For example, an operator often feels missing some views and context of the local environment, and distracts the attention due to frequently switching the two video windows. Fortunately, there is a great deal of overlap between two live videos owing to the wide-angle lens of cameras. Therefore, we propose to stitch the two live videos from the FF-camera and the DF-camera into one stitched video, and we aim at providing more compact and efficient visual feedback for the user interface of telepresence robots with user-friendly interactive experiences.

Video stitching has been widely used in video surveillance \cite{Meng2015SkyStitch}, %he2015parallax
virtual reality (VR) \cite{lee2017high}, %anderson2016jump
and digital entertainment \cite{silva2016real}. %el2009stitching,
Existing methods mainly address two challenging issues: very high computational cost and visual artifacts (e.g., jitters, causing by the lack of a spatial and temporal coherence stitching model between successive frames). In our robotic telepresence application scenario, the FF-camera uses a wide-angle lens for video communication, and the DF-camera uses a fisheye lens to obtain a full view of the telepresence robot and its surroundings for navigation. So we have to face the strong distortion of fisheye videos and non-ideal inputs (e.g., the optical centers of the cameras are not exactly at the same location, the scene is non-planar, and/or dominant foreground objects move across cameras).

A straightforward scheme to handle these challenging issues is to perform image stitching on each pair of video images. There are some works on stitching wide-angle images \cite{byrod2009minimal,ju2014stitching} and fisheye images \cite{ho2017dual,xu2017wide}, which stitch the distorted images by correcting the distortion or unwarping the distorted images with the equirectangular projection. However, the distortion correction and the unwarping process may produce unnatural effects on regions near image edges. And if being designed to improve stitching quality, these methods often suffer from a high computational cost. Directly employing image stitching algorithms for video stitching also introduces noticeable visual artifacts (e.g., jitters).

In this paper, we develop and implement a multi-homography-based video stitching algorithm to create the stitched video from a wide-angle camera and a fisheye camera for telepresence robots. Our method consists of video image alignment, seam cutting, and video image blending. To provide a visual feedback without shape distortion caused by stitching, we directly align the wide-angle video image and the fisheye video image based on the multi-homography alignment without distortion correction, unwarping with equirectangular projection, or other pre-processes. To alleviate ghosting effects caused by moving objects and/or moving cameras during telepresence robots driving, we use an enhanced dynamic programming algorithm to find an optimal seam for warped video image composition. The selected optimal seam will be updated in subsequent video images, considering spatial and temporal coherence. The final stitched video is created through image blending on the basis of the optimal seam.
We conducted a user study on a telepresence robot equipped with a wide-angle lens camera and a fisheye lens camera to demonstrate the effectiveness of our method.

Our contributions are three-fold:
\begin{itemize}
  \item We propose to stitch videos captured from a FF-camera and a DF-camera to provide more compact and efficient visual feedback for the user interface of telepresence robots, and provide the users with user-friendly interactive experiences.
  \item We present a multi-homography-based video stitching method to stitch videos from a wide-angle camera and a fisheye camera. Without calibration, distortion correction, and unwarping procedures, we can obtain a stitched video with shape preservation in the non-overlapping regions and alignment in the overlapping area.
  \item The user study results demonstrate the effectiveness of our method and the superiority of the telepresence robots with a stitched video as visual feedback.
\end{itemize}

The remainder of this paper is organized as follows. Section \ref{sec: related work} briefly reviews the related work.
Section \ref{sec: video stitching} describes the multi-homography-based video stitching algorithm for telepresence. The user study settings, procedures, results, and corresponding analyses are discussed in Section \ref{sec: user study}, and we conclude this work in Section \ref{sec: conclusions}.

\section{Related Work}
\label{sec: related work}
In this section, we make a review on image alignment and video stitching.

\subsection{Image Alignment}
Image alignment is essential for video stitching, and has attracted a lot of attentions in the past decades \cite{szeliski2006image}. Conventional methods typically estimate a global transformation to bring an image pair into alignment \cite{szeliski1997creating, brown2007automatic, li2016siasm}, making an assumption that the scene is a roughly planar, or images are captured by purely rotating the camera about its optical center. Such imaging conditions are rarely met in practice, resulting in misalignments and ghosting effects in alignment results.

To address these problems, many efforts have been devoted to estimating multiple transformation. Gao et al. \cite{gao2011constructing} estimated dual-homography for the image alignment when the scene can be divided into a distant plane and a ground plane. Lin et al. \cite{lin2011smoothly} proposed a smoothly varying affine transformation, according to the smoothly varying depth of the scene. Similarly, Zaragoza et al. \cite{zaragoza2013projective} presented an as-projective-as-possible method (APAP) to estimate multiple homographies for better alignment. Lou and Gevers \cite{lou2014image} described a piecewise planar region matching method to calculate multiple affine transformations, and they used multiple planes to approximate the image. These methods improve the alignment quality but heavily depend on keypoint detection and feature matching algorithms to offer sufficient and uniformly distributed keypoint correspondences. Additionally, keypoint detection, feature matching, and transformation estimation are time-consuming for real-time applications.

More recently, deep convolutional neural networks have been exploited to handle the problems of low efficiency and sparse keypoint correspondences for image alignment. DeTone et al. \cite{detone2016deep} designed a HomographyNet to directly estimate a homography between two images in an end-to-end manner. With the success of the HomographyNet, several deep learning-based homography estimation networks have been presented. Nowruzi et al. \cite{nowruzi2017homography} proposed a hierarchy network using a twin convolutional network, while Chang et al. \cite{chang2017clkn} presented a cascade Lucas-Kanade network by combining the Lucas-Kanade algorithm with the convolutional neural network. Apart from these supervised learning-based methods, an unsupervised homography estimation network was introduced by Nguyen et al. \cite{nguyen2018unsupervised} for UAV image alignment.

The alignment of wide-angle images and/or fisheye images is more challenging, and suffers from heavy radial distortion \cite{xu2017wide}. In order for bringing wide-angle images into alignment, Jin \cite{jin2008three} and Byr\"od et al. \cite{byrod2009minimal} estimated jointly the lens distortion and the alignment transformation, assuming that all images share the same distortion factors. For cameras in different radial distortions, Ju and Kang \cite{ju2013panoramic} estimated the lens distortion factor for each image, and then computed a homography for the alignment of synthetic images, whereas Kukelova et al. \cite{kukelova2015radial} estimated a homography and different distortion factors to bring images of real scene into alignment. In addition, Ho and Budagavi \cite{ho2017dual} proposed to align two images captured by a dual-fisheye lens camera. They unwarped the fisheye images into spherical 2-Dimensional space, and then employed a two-step alignment to register the unwarped images. Due to the unwarping process, the regions near edges of original images are stretched, leading to shape distortions in alignment results.

In our application scenario, we need to bring a wide-angle video image and a fisheye video image into alignment. To obtain an alignment with shape preservation in both the fisheye video image and the wide-angle video image for telepresence, we also need to bring the image pair into alignment without any distortion correction and unwarping processes.

\subsection{Video Stitching}
There are several commercial video stitching softwares, such as VideoStitch Studio\footnote{https://www.orah.co/software/videostitch-studio/} and AutoPano\footnote{http://www.kolor.com/autopano/}. These softwares usually compute a 2D transformation relating two cameras, and then bring all pairs of video images into alignment for the post-production of stitched videos. To improve the quality of the stitched video, Li et al. \cite{li2015efficient} found double-seam to eliminate intensity misalignment, and similar work has been presented for designing a content-aware adaptive blending \cite{kim2017content}. Some works were presented to obtain better alignment results. Lee and Sim \cite{lee2015robust} stitched videos by projecting the background plane and the foreground objects separately, while Jiang and Gu \cite{jiang2015video} stitched videos using spatial-temporal content-preserving warps. For videos captured by handheld cameras, Su et al. \cite{Su2016Video, nie2018dynamic} and Guo et al. \cite{Guo2016Joint} combined the stitching and stabilization techniques together into a unified optimization framework for video stitching, whereas Lin et al. \cite{lin2016seamless} stitched videos by reconstructing the 3D scene using the recovered 3D camera paths and the 3D scene points. These methods stitched videos in an iterative manner with low computational efficiency.

Besides, some work was designed for real-time processing or time-critical applications. For video surveillance applications, He and Yu \cite{he2015parallax} employed a background modeling algorithm and a change-detection-based optimal seam selection approach to stitch videos captured by fixed cameras. A Multi-UAV-based video surveillance system, SkyStitch \cite{Meng2015SkyStitch}, was designed and implemented for real-time aerial surveillance, employing flight information (e.g., the UAV attitude and GPS location) got from the flight controller as assistance. Okumura et al. \cite{okumura2013real} introduced a real-time video stitching method by implementing and improving a feature-based algorithm on a field-programable gate array (FPGA). Apart from hardware acceleration, EI-Saban et al. \cite{el2009stitching} developed a real-time method to stitch independent videos streamed by different mobile phones, while Silva et al. \cite{silva2016real} stitched several live videos into a $360^o$ field of view and spread the stitched video based on GPU.

Since most existing approaches are designed for stitching videos from conventional cameras, they can not handle videos with heavy lens distortions captured by the wide-angle lens camera or fisheye lens camera. Considering the distortion, a simple method is to undistort the video images through a rectilinear projection, and then stitch the undistorted videos frame-by-frame \cite{perazzi2015panoramic}. Nevertheless, the undistortion may incur unnatural stretches on the regions near the borders of video images, particularly in video images captured by fisheye lens cameras.

Different from the existing methods, we stitch two live videos captured by a wide-angle lens camera and a fisheye lens camera mounted on a telepresence robot, and provide more compact and efficient visual feedback for the users to obtain friendly interactive experience. Without calibration, distortion correction, and unwarping procedures, we can obtain a stitched video with shape preservation in the non-overlapping regions and alignment in the overlapping area for telepresence.

\begin{figure*}
  \centering
  \centerline{\includegraphics[height=9.0cm]{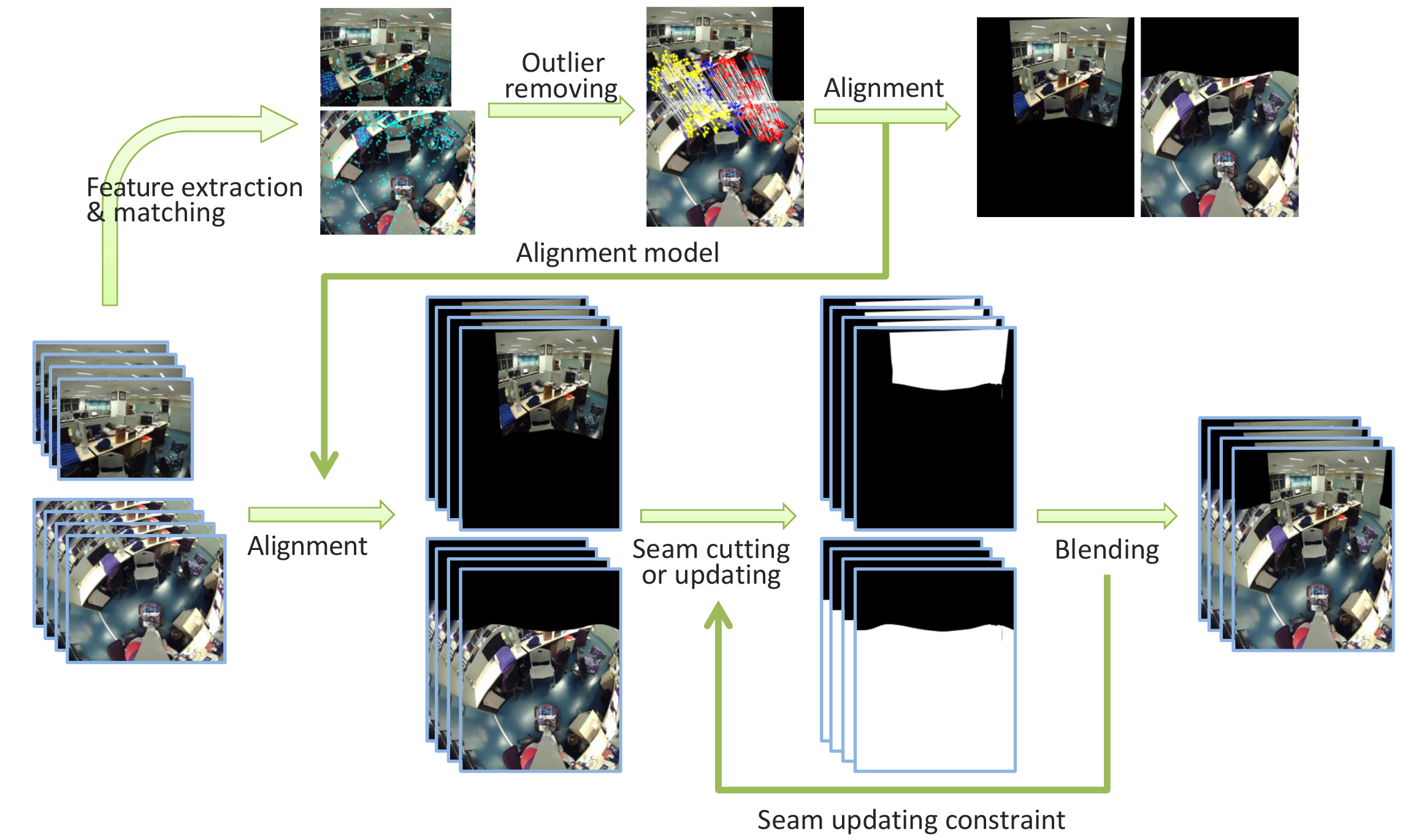}}
\caption{Pipeline of the video stitching of a wide-angle video and a fisheye video.}
\label{fig: video stitching}
\end{figure*}

\section{Video Stitching for Telepresence}% Real-Time Video Stitching for Telepresence
\label{sec: video stitching}
In this section, we describe the algorithm that used to stitch the videos captured from a wide-angle camera and a fisheye camera mounted on a telepresence. We combine video image alignment, seam cutting and updating, image blending together to stitch the videos for telepresence, considering spatial and temporal coherence to alleviate the jitters. The pipeline of the video stitching algorithm is depicted in Fig. \ref{fig: video stitching}.

\subsection{Video Image Alignment}
\label{sec: alignment}
Without calibration, distortion correction, and unwarping procedures, we align the wide-angle video image and the fisheye video image using the multi-homography alignment method proposed in our previous work \cite{xu2017wide}. A keypoint detector and descriptor (e.g., SUFT \cite{Bay2006SURF}) is used to obtain feature points from the two video images. After feature matching, inliers can be selected from the point correspondences by using a multi-homography inlier selection method.
A global projective transformation and multiple local homographies are then estimated from the inliers. The final multi-homography warps are constructed by weighting between the global homoggraphy and local homography, in which local homographies are exploited for local region alignment. As a result, we can achieve a good alignment accuracy in the overlapping area and shape preservation in non-overlapping regions.

\subsubsection{Multi-Homography Inlier Selection}
\label{subsec: inlier_selection}
Feature matching can produce point correspondences from all the feature points, and there may include many mismatched points (i.e., outliers). To remove outliers from the point correspondence set, an inlier selection method can be employed. The Random Sample Consensus (RANSAC) \cite{fischler1981random} is popularly used to select inliers by generating multiple hypotheses for homography estimation.

In our work, the heavy lens distortion of the wide-angle video image $I$ and the fisheye video image $I'$ should be taken into consideration, since the homography is a plane transformation. We employ a multi-homography inlier selection method which can select more inliers for the alignment. A conditional sampling strategy \cite{chin2012accelerated} is used to generate multiple homography hypotheses.
Given a set of point correspondences $F=\{(f_i, f_i')\}_{i=1}^{\hat{N}}$, we generate $M$ homography hypotheses $\{h_1,\ldots, h_M\}$ by randomly sampling $M$  minimal subsets of point correspondences from $F$, where $(f_i,f_i')$  (denoted as $F_i$) is a point correspondence between the wide-angle video image and the fisheye video image. For each point correspondence $F_i$, its corresponding residuals to all homography hypotheses are calculated and ranked in a nondescending order. According to the residual order, a new list of homography hypotheses for $F_i$ can be acquired by $h^i = \{h_1^i,\ldots, h_M^i\}$. $F_i$ is more likely to be an inlier of the hypothesis with a lower residual.

There may be many common hypotheses shared by $h^i$ and $h^j$ at the top place of the list, especially when $F_i$ and $F_j$ are in the same local area.
A conditional inlier probability is computed to guide the inlier selection after selecting $F_i$ as an inlier:
\begin{equation}
\label{eq: probability}
    f(F_i, F_j) = \frac{1}{m}|h_{1:m}^i \cap h_{1:m}^j|,
\end{equation}
where $h_{1:m}$ specifies the first-$m$ hypotheses in the list, $|\cdot|$ is the counting operator, and $\cap$ is the intersection operator. Given the first selected inlier, the probability of that inlier with the rest point correspondences of $F$ is used to select the second inlier.

In the experiment, the size of the minimal subset $s$, the outlier threshold $\varepsilon_o$ and $\varepsilon_r$, the number of homography hypotheses $M_0$ and $M$ are set to 4, 1, 0.01, 10, and 500, respectively.

\subsubsection{Global Homography Estimation}
\label{subsec: global_trans}
Given an inlier set $P = \{p_i,p_i'\}_{i=1}^N$ between $I$ and $I'$, the global homography $H_g\in \mathbb{R}^{3 \times 3}$ is defined by
\begin{equation}
\label{eq: definition}
p'\sim H_g p,
\end{equation}
where $\sim$ denotes an equality up to a scale. $p$ and $p'$ are represented in 2D homogenous coordinates, and $H_g$ is also in homogeneous. Omitting the scale term, Eq.(\ref{eq: definition}) can be rewritten as
\begin{equation}
\label{eq: global}
    \left[ \begin{array}{c} x' \\ y' \\ 1 \end{array} \right]
    =
    \left[ \begin{array}{ccc}
    h_1 & h_2 & h_3\\ h_4 & h_5 & h_6\\ h_7 & h_8 & h_9
    \end{array} \right]
    \left[ \begin{array}{c} x \\ y \\ 1 \end{array} \right].
\end{equation}

Through a cross product on both side, Eq.(\ref{eq: global}) becomes $\mathbf{0}_{3\times1} = p'\times H_g p$, and can be linearized as
\begin{equation}
\label{eq: cross product}
    \mathbf{0}_{3\times1} = \left[
    \begin{array}{ccc}
      \mathbf{0}_{1\times3}  &        -p^{\top}        &   y'p^{\top} \\
         p^\top              & \mathbf{0}_{1\times3}   &   -x'p^\top \\
         -y'p^\top           &         x'p\top         & \mathbf{0}_{1\times3} \\
    \end{array}
    \right]
    [h_1,\ldots, h_9]^{\top} = A\mathbf{h},
\end{equation}
and only two rows of $A$ are linearly independent. Using the Direct Linear Transformation (DLT) \cite{hartley2003multiple}, a global homography to fit all inliers can be solved by
\begin{gather}
\label{eq: H function}
    \mathbf{h^*} = \mathop{\arg\min}_{\mathbf{h}} \sum_{i=1}^N \|\tilde{A}_i\mathbf{h}\|^2 = \mathop{\arg\min}_{\mathbf{h}}\|\tilde{A}\mathbf{h}\|^2, \\
    s.t.\quad \quad \|\mathbf{h}\|^2 = 1 \nonumber,
\end{gather}
where $\tilde{A}_i$ is the first two rows of $A$ for the $i$-th inlier, and $\tilde{A}\in\mathbb{R}^{2N\times9}$ is the stack of all $\tilde{A}_i$.

Eq.(\ref{eq: H function}) can be solved through singular value decomposition (SVD), and global homography $H_g$ is obtained by reshaping $\mathbf{h^*}$ into a $3\times 3$ matrix.

\subsubsection{Multi-Homography Estimation}
\label{subsec: multi-h}
Due to strong distortions, wide-angle video image and fisheye video image alignment with a global homograhy will introduce large misalignment \cite{xu2017wide}. To increase the alignment quality, multiple local homographies are estimated by performing Moving DLT \cite{zaragoza2013projective} on the inlier set $P$ through
\begin{gather}
\label{eq: local_h}
  \mathbf{h}^* = \mathop{\arg\min}_{\mathbf{h}} \sum_{i=1}^N \|w_i \tilde{A}_i\mathbf{h}\|^2 = \mathop{\arg\min}_{\mathbf{h}} \|W\tilde{A}\mathbf{h}\|^2,  \\
  s.t.\quad \|\mathbf{h}\|^2 = 1, \nonumber
\end{gather}
for each position $p^*$ in image $I$, and $W\in\mathbb{R}^{2N\times2N}$ takes the form as $W=diag\left(\left[w_1,w_1,\ldots, w_N,w_N\right]\right)$. The scalar weight $w_i$ is defined as
\begin{equation}
\label{eq: weights}
  w_i = \max \left( \exp\left(\frac{-\|p^*-p_i\|^2 }{\sigma^2}\right), \gamma\right),
\end{equation}
where $\sigma$ is a scale parameter, and $\gamma \in [0,1]$ is used to avoid numerical issues. The inlier closer to $p^*$ is given a higher weight, assuming that pixels in a local area share a homography.

Due to a lack of point correspondences, local homographies in non-overlapping regions are also calculated by inliers (i.e., some point correspondences in the overlapping region). To alleviate the artifact, we integrate the local homography $\mathbf{h}^*$ and the global homography $H_g$ into a new homography, taking advantage of both homographies for local alignment and shape preservation. The integration formulation is given by
\begin{equation}
\label{eq: multi_h}
  H = w H_l + (1-w)H_g,
\end{equation}
where $w$ is the integration parameter, and $H_l$ is the $3\times3$ matrix form of the local homography $\mathbf{h}^*$.

Since the deformation caused by the global homography increases along the positive $u$-axis from the overlapping region to the non-overlapping regions \cite{chang2014shape}, $H$ is smoothed from local homography to global homography along $u$-axis using $w$,
\begin{equation}
\label{eq: h_coefficient}
  w = (u - u_m)/(u_M - u_m),
\end{equation}
where ($u,v$) is a new coordinate obtained by rotating the original coordinate ($x,y$) of the warped image of $I$.
$u_m$ and $u_M$ are the minimum and the maximum $u$ coordinate of all pixels, respectively. The rotation angel is $\theta=\arctan(h_8/h_7)$. Note that image $I'$ is warped by using $R=H(H_l)^{-1}$ to compensate the local homography warping effects on the overlapping region.

\subsection{Optimal Seam Cutting and Updating}
\label{subsec: seam cutting}
To alleviate ghosting effects caused by moving objects and/or moving cameras, an optimal seam is found for aligned video composition, and the optimal seam will be updated in subsequent frames, considering spatial and temporal coherence. The final stitched video image, called forward-downward-facing video image (FDF-video image), is created by using the multi-band blending to provide a smooth transition of lighting from the downward-facing video image (DF-video image) to the forward-facing video image (FF-video image).

\subsubsection{Seam Cutting}
\begin{figure}[t]
  \centering
  \centerline{\includegraphics[width=8.0cm]{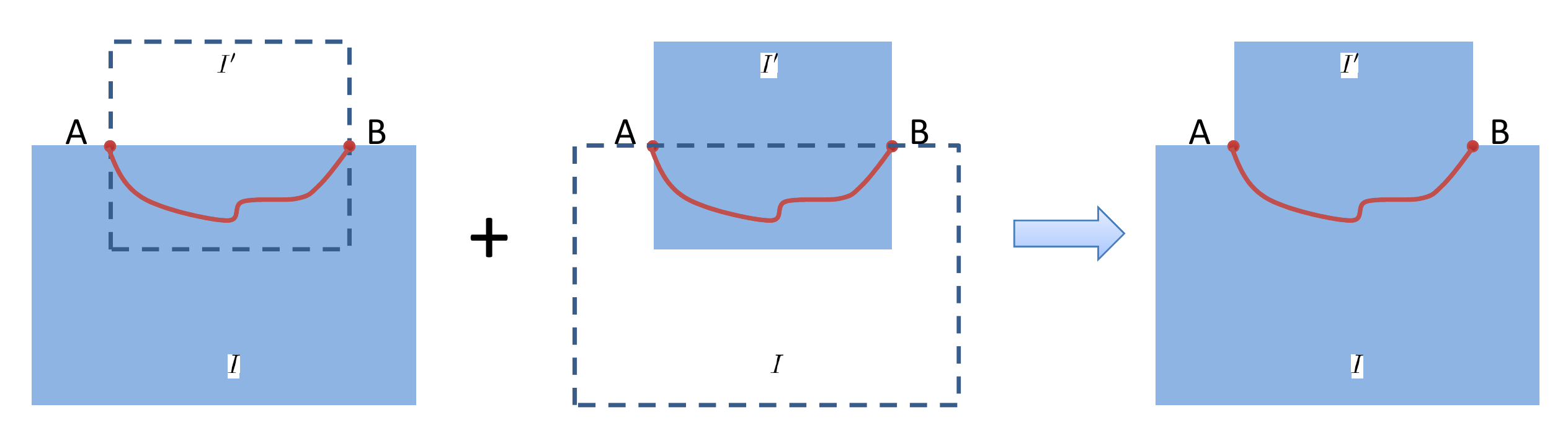}}
\caption{Seam Cutting. The task is to find an optimal seam from point A to point B, such as the red line.}
\label{fig: seam_cutting}
\end{figure}

\begin{figure}[t]
  \centering
  \centerline{\includegraphics[width=2cm]{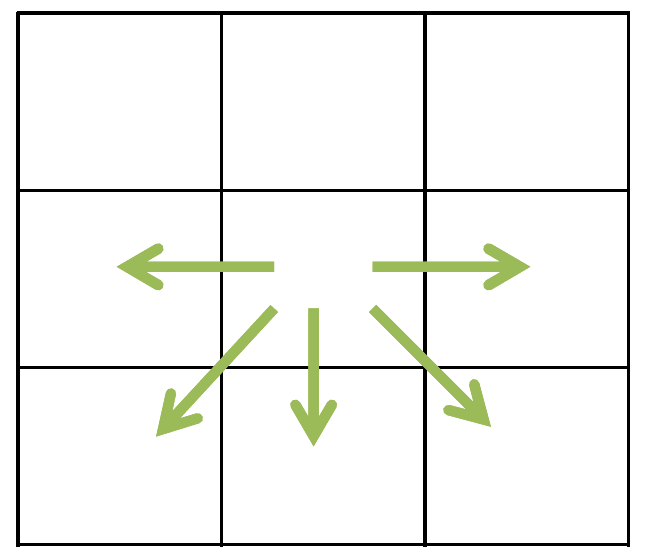}}
\caption{Search directions of the enhanced dynamic programming algorithm.}
\label{fig: search_direction}
\end{figure}

Seam cutting is utilized to select an optimal pixel-based continuous curve (seam) for image blending, which can alleviate ghosting effects caused by moving objects and/or moving cameras. As demonstrated in Fig. \ref{fig: seam_cutting}, the seam cutting task is to find an optimal seam, e.g., the red line from point A to point B over the overlapping area. We use an enhanced dynamic programming approach \cite{gu2009new} that holds search directions as shown in Fig. \ref{fig: search_direction} to find the optimal seam.

The enhanced seam is defined as
\begin{multline}
\label{eq: seam definition}
   C_{i,j} = e_{i,j} + {}\\ \min\left( C_{i-1,j-1}, C_{i-1,j}, C_{i-1,j+1}, C_{i,j-1}, C_{i,j+1} \right),
\end{multline}
where $(i,j)$ is a pixel coordinate, $C$ and $e$ indicate the cumulative cost and the gradient cost, respectively. To find a seam without gradient difference and visible artifacts, we define the gradient cost by gradient smoothness $S_m$ and gradient similarity $S_d$:
\begin{equation}
\label{eq: cost value}
e = S_m + S_d.
\end{equation}
Given the overlapping areas $I_s$ and $I_t$ of the two warped images, $S_m$ and $S_d$ are defined as
\begin{gather}
\label{eq: gradient smoothness}
S_m = \|\nabla(I_s + I_t)\| / mean{\left(\|\nabla(I_s+I_t)\|\right)}, \\
S_d = \|\nabla(I_s - I_t)\| / mean{\left(\|\nabla(I_s-I_t)\|\right)},
\end{gather}
where $\|\cdot\|$ and $\nabla$ are the L2-norm and the gradient operator, respectively.

\subsubsection{Selected Seam Updating}
The selected seam will be dynamically updated in successive video images. Avoiding to introduce noticeable artifacts caused by large drift of the optimal seams between successive video images, we employ a seam updating method based on temporal propagation constraint \cite{peng2016fast} to gain stable seams.

The temporal propagation constraint is constructed based on location information of the optimal seam in the previous video image. It is represented by a matrix $\mathbf{D}^{t-1}_{w\times h}$, where $t$ is the index of the video image. Each element of $\mathbf{D}^{t-1}_{w\times h}$ is a penalty for each point in the overlapping area, which equals to horizontal distance between the corresponding point and the optimal seam of the previous video image. In other words, the penalty increases with the distance.

For the current video image, we can get a cost matrix $\mathbf{C}^t_{w\times h}$ using Eq. (\ref{eq: seam definition}). Combining with the temporal propagation constraint, the final cost matrix $\mathbf{\tilde{C}}^t_{w\times h}$ is calculated using
\begin{equation}
 \mathbf{\tilde{C}}^t_{w\times h} = \mathbf{C}^t_{w\times h} + \mathbf{D}^{t-1}_{w\times h}.
 \end{equation}
We can update the selected seam by performing the enhanced dynamic programming algorithm again.

\begin{figure}[t]
\subfigure[]{
\centering
  \begin{minipage}{0.95\linewidth}
  \centering
  \centerline{\includegraphics[width=8.6cm]{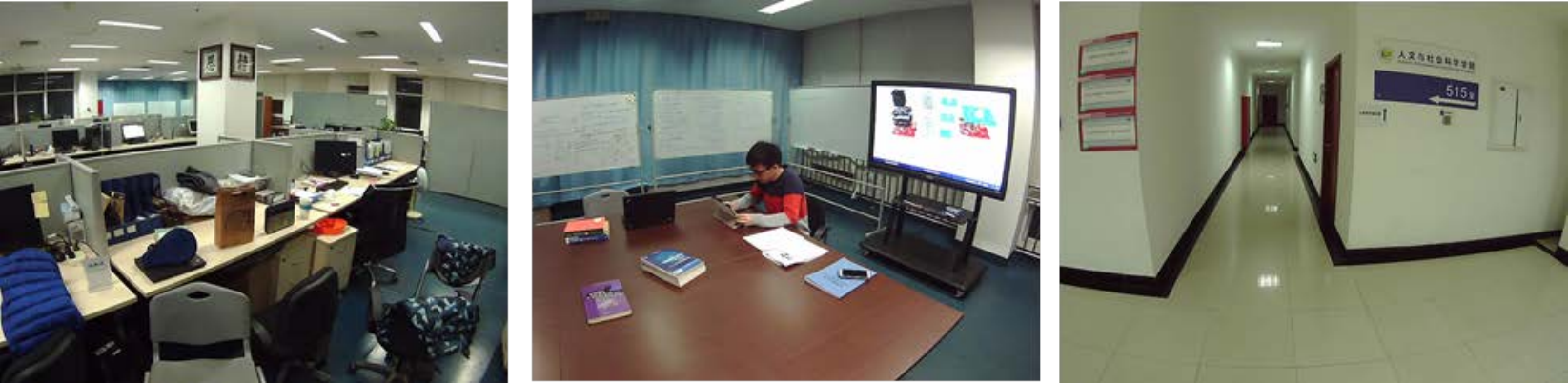}}
  \end{minipage}
}
\subfigure[]{
\centering
  \begin{minipage}{0.95\linewidth}
  \centering
  \centerline{\includegraphics[width=8.6cm]{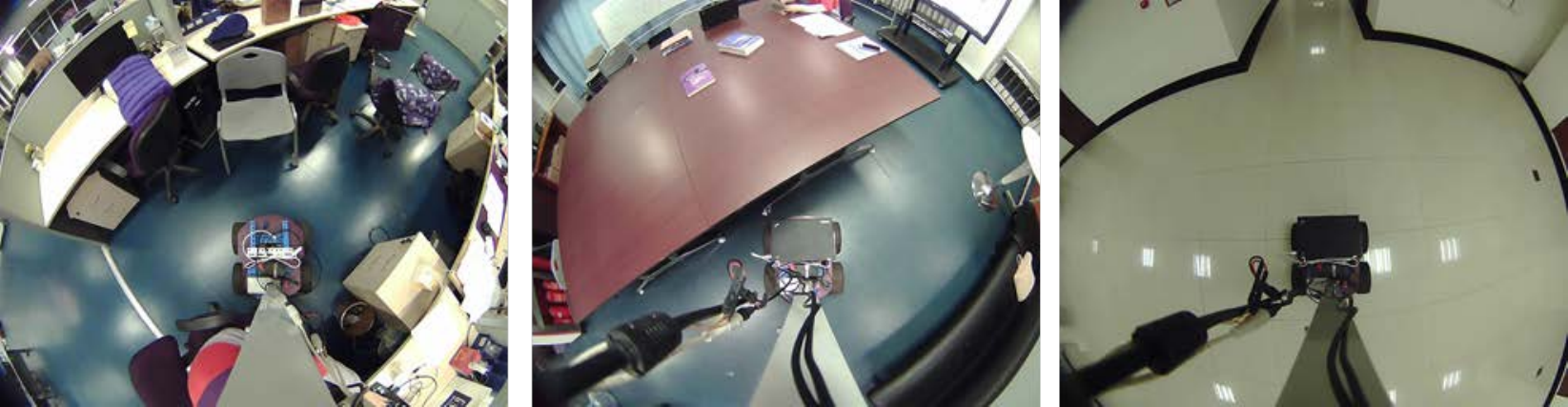}}
  \end{minipage}
}
\subfigure[]{
\centering
  \begin{minipage}{0.95\linewidth}
  \centering
  \centerline{\includegraphics[width=8.8cm]{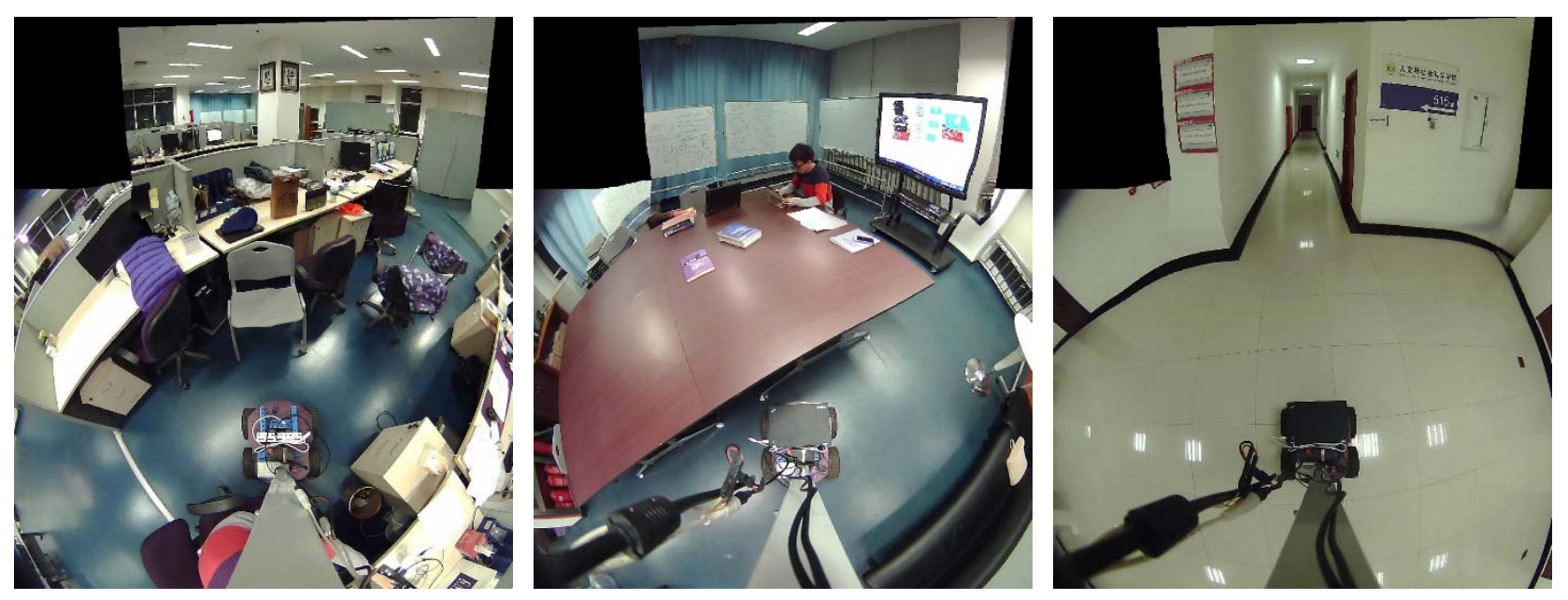}}
  \end{minipage}
}
\caption{Stitching results on wide-angle images and fisheye images of different scenes. (a) Original wide-angle video images captured by the FF-camera. (b) Original fisheye video images captured by the DF-camera, corresponding to the original wide-angle video images in (a). (c) Stitching results (the FDF-video images). }
\label{fig: image_stitching}
\end{figure}

\subsection{Video Image Blending}
Due to the distinct orientations of the DF-camera and the FF-camera, there exist lighting inconsistencies between the same scenes of the DF-video image and the FF-video image. Video image blending can be used to achieve a smooth transition of lighting from one image to the other. We utilize the multi-band blending \cite{burt1983a} which is widely used and relatively insensitive to misalignment \cite{zhu2016comparative}, for warped video image composition.

We build a Laplacian pyramid on each warped video image, and the blending becomes a solution of feather blending on each pyramid level. To obtain the weights used to perform feather blending, each weight image built on the optimal seam is converted into a Gaussian pyramid, and each Gaussian pyramid level is a weight map for corresponding level of the Laplacian pyramid. The FDF-video image is reconstructed by interpolating and merging all the blended pyramid levels.

\subsection{Stitching results}
\label{subsec: experiments}
In general, each image pair should be aligned using the alignment algorithm. However, alignment on every newcome image pair will make the stitching slow due to the low computational efficiency of keypoint detection, feature matching, and inlier selection. Fortunately, the DF-camera and the FF-camera are fixed on the vertical post of the telepresence robot. We can assume that the configuration of the cameras are not changed in a matter of seconds, so we do not need to estimate the alignment model on every image pair.

Some stitched video images and the corresponding original wide-angle video images captured by the FF-camera and fisheye video images captured by the DF-camera are shown in Fig. \ref{fig: image_stitching}. These images consists of complex scenes and simple scenes, and are captured by a telepresence robot with a wide-angle lens camera and a fisheye lens camera.

\section{User Study}
\label{sec: user study}
We conducted a user study to compare visual feedback of a stitched video (the FDF-video) with traditional two separate videos (the DF-video and the FF-video) for telepresence robots.

\subsection{Experimental Platform}
The experimental platform consists of a telepresence robot in a local environment, user interfaces used by an operator in a remote environment, and wireless communication networks for connecting the two environments. An operator can use a smart pad in a remote environment to drive a telepresence robot to acquire the live video of a local environment as visual feedback.

\begin{figure}[t]
  %\centering
  \centerline{\includegraphics[width=5cm]{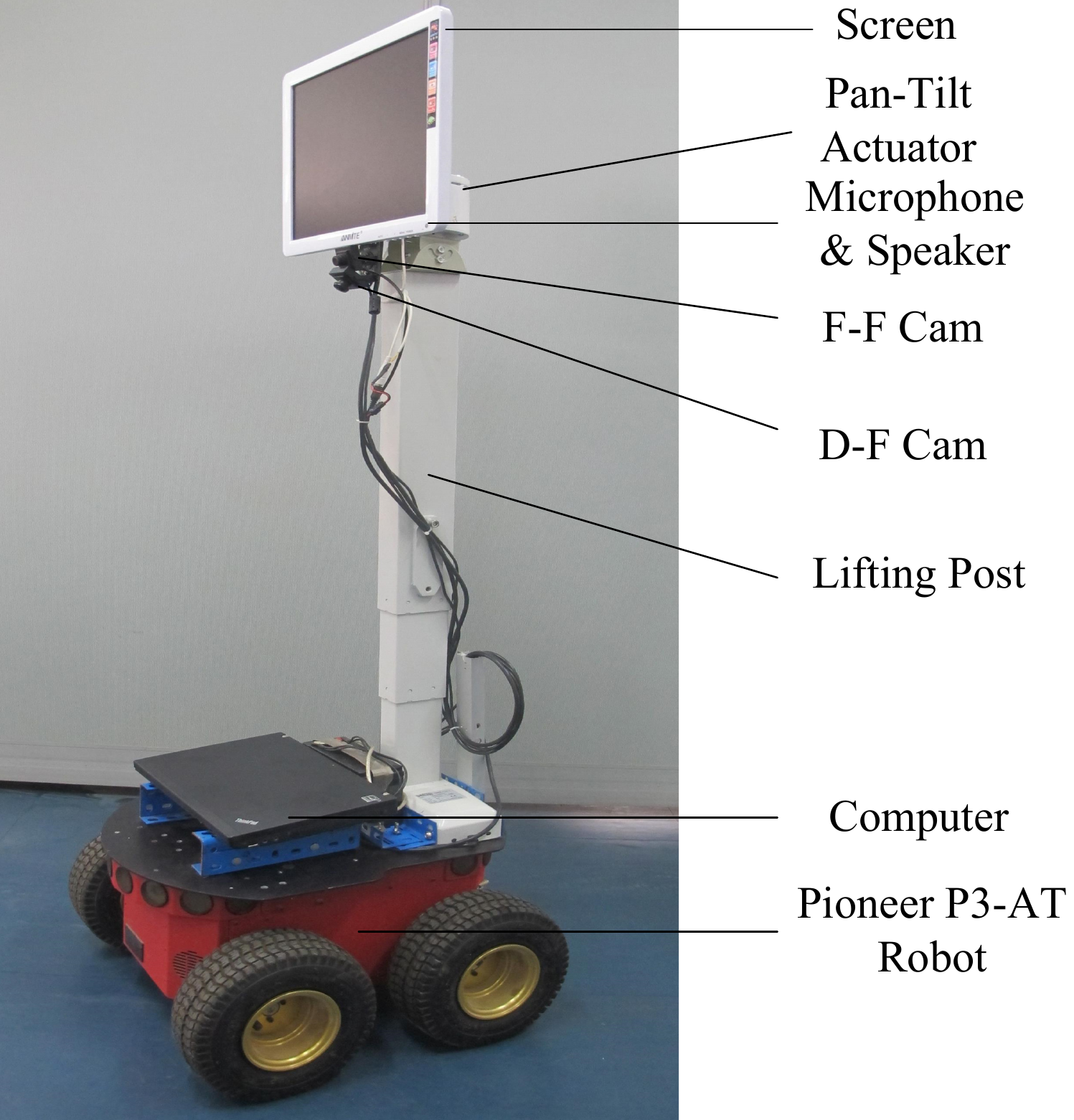}}
\caption{The Mcisbot robot.}
\label{fig: mcisbot}
\end{figure}

The telepresence robot used in our user study was developed in our lab, called Mcisbot \cite{jia2015telepresence}, as depicted in Fig. \ref{fig: mcisbot}. It uses the Pioneer 3-AT as a mobile robot base equipped with a special designed robot head for telepresence. The robot head contains a light LCD screen, a forward-facing camera (FF-camera), a downward-facing camera (DF-camera), and a speaker $\&$ microphone, and all together are mounted on a pan-tilt platform hold up by a vertical post. The FF-camera with a wide-angle lens can provide a live video for clear watching of targets or persons in front. The DF-camera with a fisheye lens provides a complete watching of the ground around the robot for navigation.

\begin{figure}[t]
\centering
\subfigure[TVUIs]{
  \begin{minipage}[t]{0.9\linewidth}
    \centering
    \centerline{\includegraphics[width=8.cm]{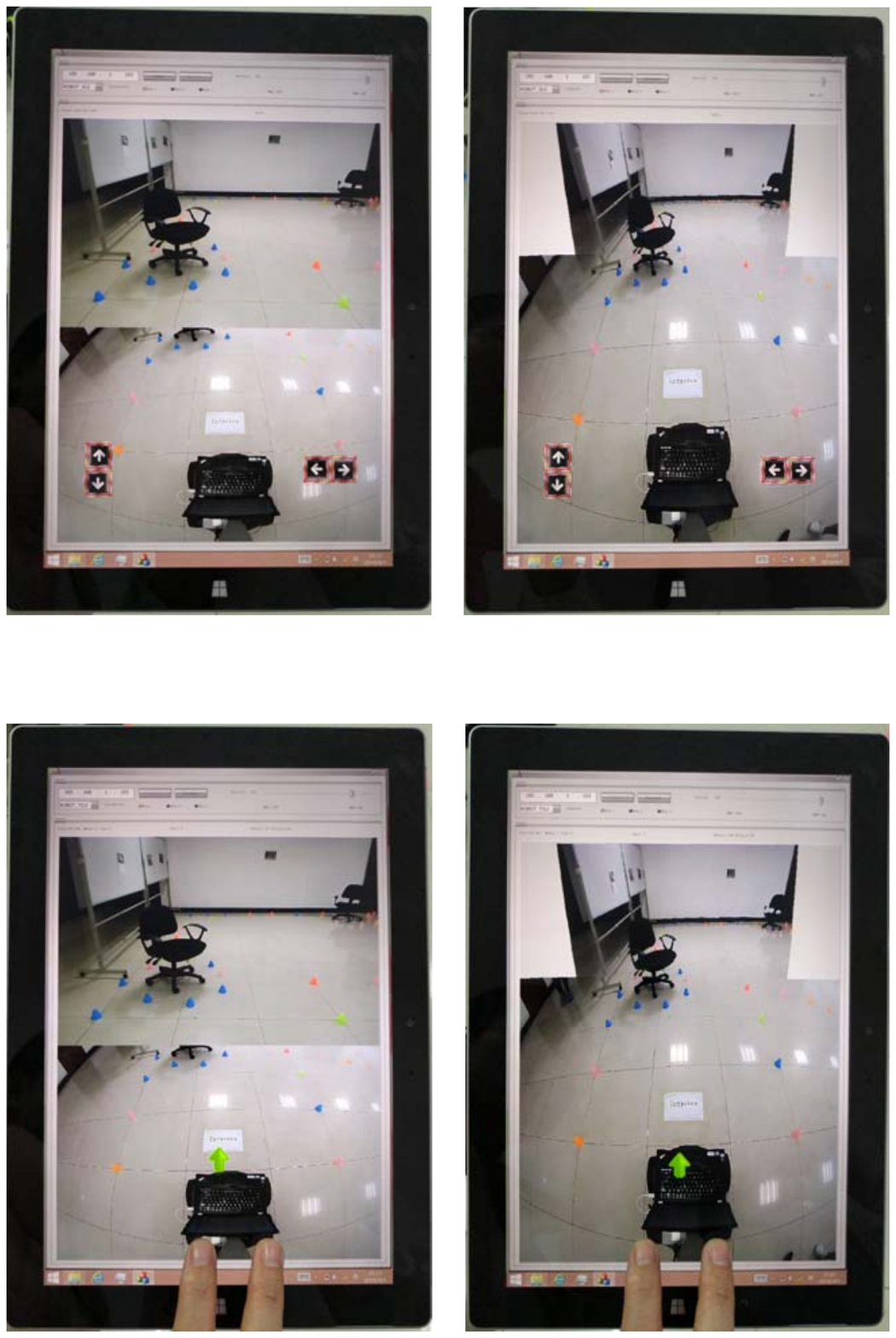}}
  \end{minipage}
}
\subfigure[GUIs]{
  \begin{minipage}[t]{.9\linewidth}
    \centering
    \centerline{\includegraphics[width=8.cm]{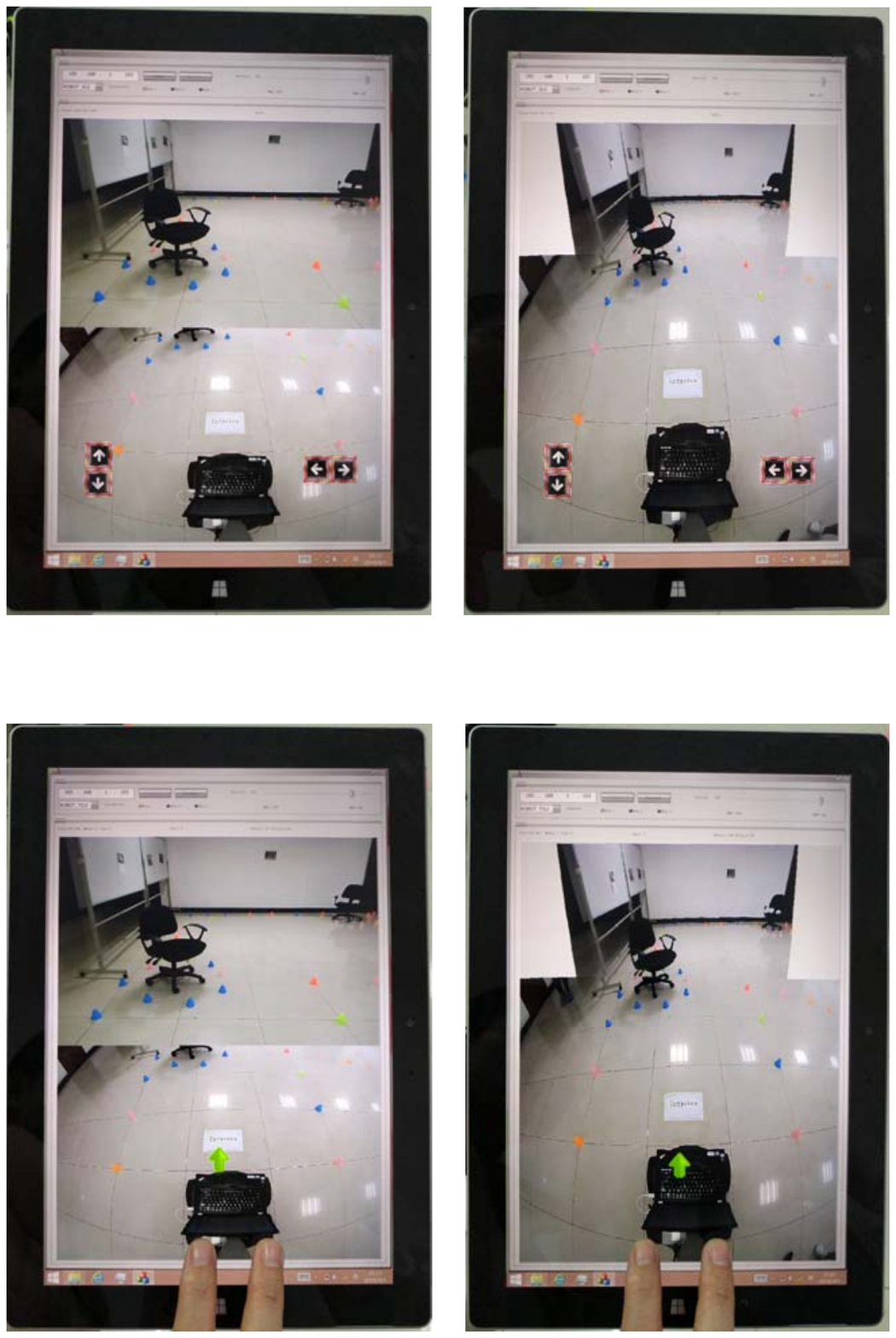}}
  \end{minipage}
}
\caption{Two kinds of user interfaces used in our user study. (a) The GUI with two separate videos (left) and the GUI with a FDF-video (right). (b) The TVUI with two separate videos (left) and the TVUI with a FDF-video (right). }
\label{fig: TIUI&GUI}
\end{figure}

The Mcisbot was specifically designed to evaluate the usability of the Touchable live Video image based User Interface (TVUI). The most notable feature of the TVUI is that there are no explicit graphical buttons, arrow keys, and/or menus, compared to traditional touchscreen GUIs. The TVUI allows operators to drive the robot by directly touching the live video images with finger touch gestures. Naturally, we use the TVUI to test the effectiveness of the stitched video for convenience. Moreover, since traditional touchscreen GUIs are widely used to support pilot operators to teleoperate telepresence robots \cite{Double2017,Beam2017, tsui2014iterative,mayer2014user}, we also conducted the user study on the GUIs to evaluate the effectiveness of the proposed method.
Fig. \ref{fig: TIUI&GUI}~(a) shows the TVUI containing two separate videos (the FF-video and the DF-video) and the one with a stitched video (the FDF-video). The GUIs used in our experiments are shown in Fig.~\ref{fig: TIUI&GUI}~(b).

During the experiment, we resized both the wide-angle video image and the fisheye video image to 640$\times$480 for stitching. Using a laptop computer with an I5-2410M Intel 2.30 GHz CPU and 4 GB RAM, we achieved a rate of video stitching up to 15 fps (frames per second). Considering system delay and video stitching efficiency, we directly show the stitching result in the user interfaces as visual feedback without any post-processing.

\subsection{Participants}
We recruited $18$ participants from the local university for the user study, whose ages vary from $17$ to $28$ years (M=21.28, SD=2.803), where M and SD indicate the mean value and the standard deviation, respectively. All participants use computers in their daily life. With a five-point scale for familiarity, ranging from ``1 = not at all familiar'' to ``5 = very familiar'', all participants reported their familiarity with telepresence robots (M=1.72, SD=0.752). A few of them heard of telepresence robots, but had no experience in telepresence robot operation. The others expressed that they had no idea about telepresence robots. Similarly, all the participants reported how familiar they were with video chat on the same five-point scale, and most of them had video chat experiences (M=3.83, SD=1.200).

\subsection{Environment Setup}
\begin{figure}[t]
\centering
\subfigure[]{
    \begin{minipage}[t]{0.9\linewidth}
    \centering
    \centerline{\includegraphics[width=9cm]{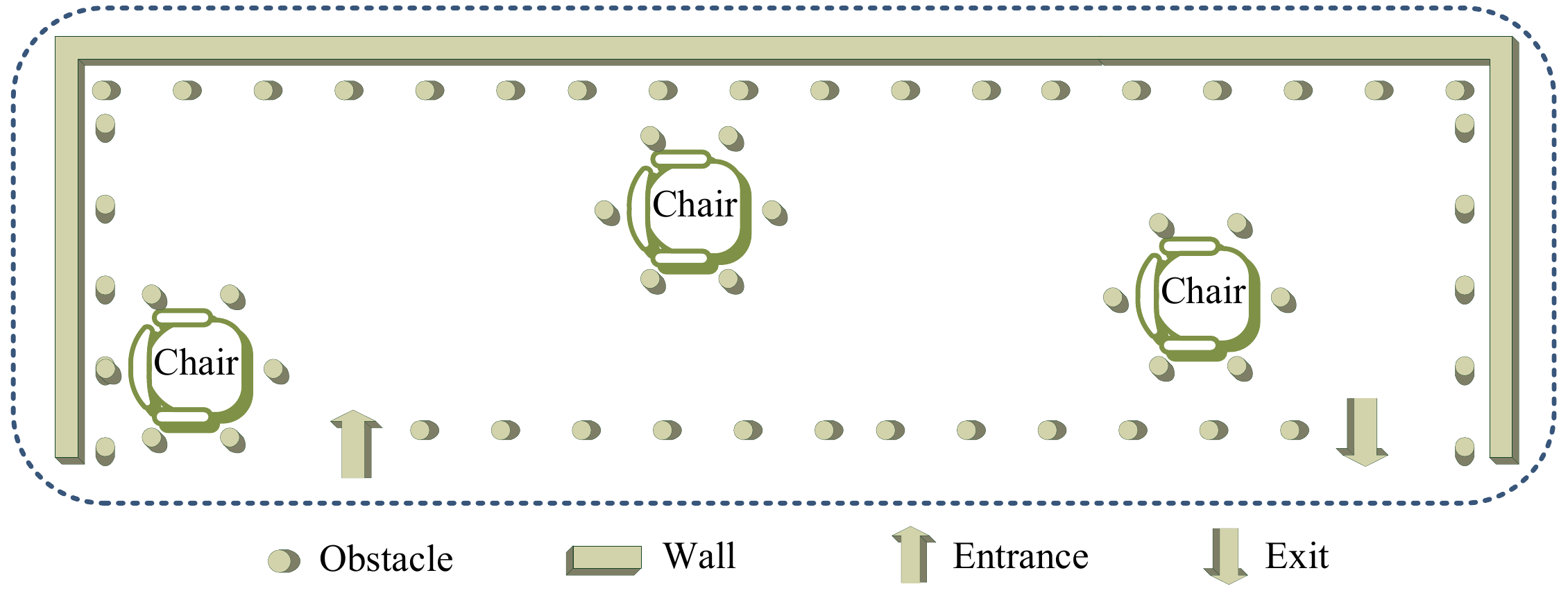}}
    \end{minipage}
}
\subfigure[]{
    \begin{minipage}[t]{0.9\linewidth}
    \centering
    \centerline{\includegraphics[width=8.8cm]{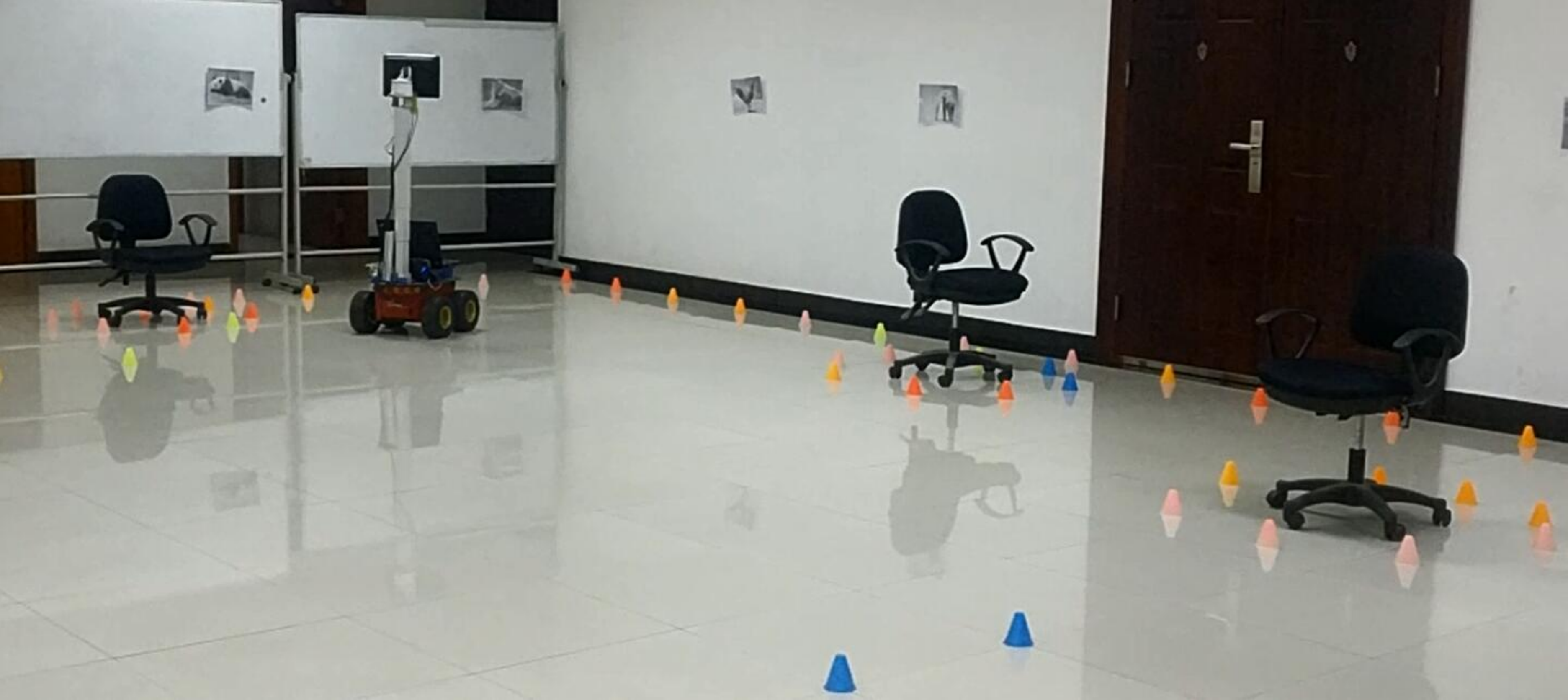}}
    \end{minipage}
}
\caption{The local environment. (a) The physical arrangement of the local environment. (b) The robot was driven to walk around the local environment. }
\label{fig: environment}
\end{figure}

We constructed an experimental room in our lab as a local environment to simulate a complex environment, such as a museum or a meeting room that contains some obstacles, pictures, and chairs. The physical arrangement of the local environment is shown in Fig. \ref{fig: environment}(a). The obstacles and chairs offered participants the direction to drive the robot to walk around the local environment, and also played roles as anti-collision objects for safely driving. Operators were located in another room, specified as a remote environment, to remotely drive the robot using the user interface, and a picture of an operator driving the robot to walk around the experimental room is shown in Fig. \ref{fig: environment}(b).

\subsection{Tasks}
Providing video(s) as visual feedback, telepresence robots are typically used for telecommunication and teleoperation. We designed a task of animal picture recognition and robot driving for comparison between the visual feedback with a stitched video and traditional two separate videos.

For one kind of the visual feedback, a participant was required to drive the Mcisbot robot to walk through the local environment in safe. Meanwhile, he/she should recognize all the animal pictures on walls and tell the experimenter what animal he/she has saw. There were 8 animal pictures on walls and with different recognition difficulty, e.g., different contrast between the foreground and the background.

The two cameras are designed for different purposes, the DF-camera for navigation and the FF-camera for communication. To keep the robot safe from collision with other objects, operators needed to focus on the DF-video (or the lower part of the FDF-video) all the time when they were driving the robot. However, to recognize pictures on walls, participants needed to watch the FF-video (or the upper part of the FDF-video). For safely driving, participants may stop driving when they could not watch the ground around the robot.

\subsection{Measurements and Analyses}
We investigated objective and subjective measurements for visual feedback with a stitched video and two separate videos.

The objective measurements include task performance and situation awareness ability through the user interface. We measured task performance using the task completion time, timing from the robot entering the entrance to it coming out from the exit. The situation awareness ability was measured by the number of pictures that the participant had saw, and the number of correctly recognized pictures.

The subjective measurements were gained through three questionnaires, and consists of the situation awareness ability, perceived task success, and the participants' preference between the user interface with the stitched video and the traditional two separate videos. For the participants' preference, we compared the number of persons for each option. There was a five-point scale for other questions, and the larger was the better.

A one-way fixed-effects analysis of variance (ANOVA) was conducted to test the effects of the two cases upon measurements of task completion time, situation awareness ability, and perceived task success. For tests of statistical significance, we used a cut-off value of $p<0.05$.

\subsection{Procedure}
A mixed between- and within-subject user study was conducted. The user interface type (the TVUI and the GUI) was the between-subject variable, and all participants were divided into two groups, one group using the TVUI and the other group using the GUI. The visual feedback (the FDF-video and the two separate videos) was the within-subject variable such that a better comparison between the two manners can be obtained. To counterbalance the possible ordering effects, we permuted the order of visual feedback manner used between the participants. We changed the locations of animal pictures and furniture after the first trial, and told the participants that the physical layout of the environment may have been changed, such that they need to re-explore the environment to complete the task again.

In preparation, a participant was given an overview of the experimental task, and then an experimenter provided the instructions on how to use the TVUI or the GUI to drive the Mcisbot robot. All participants were given $10$ minutes to practice remotely driving the robot with the help of an experimenter in the training room. This training environment contained several different animal pictures for participants to practice the recognition task. Most of participants completed the practices in less than $10$ minutes. And most of participants spent practice time with the TVUI exceeding the GUI because the GUI is simple and intuitive while skilled operation with touchscreen gestures on the TVUI need to be obtained. But using the TVUI to remotely drive the robot is more flexible than the GUI, particularly changing the direction and speed of the robot.

After the experimenter drove the robot to entrance of the local environment, a participant started to operate through the user interface with one manner, and the timer was started until the robot came out from the exit. The number of the correctly recognized pictures and the total number of the saw pictures were accounted by the experimenter. The participant was asked to fill in the first questionnaire and prepared to using the user interface with the other manner to start the task again. Before the task, the experimenter changed the locations of the pictures and furniture, and told the participant that the layout of the environment may have been changed. Then, the participant started to re-explore the local environment again to complete the task. After the task, the participant was required to fill in the second questionnaire, and the same objective measurements were recorded. Finally, the third questionnaire designed for capturing the participant's preference on the stitched video and traditional two separate videos was also required to be filled in.

\subsection{Results}
\begin{figure}[t]
\centering
\subfigure[Objective measurements]{
    \begin{minipage}[t]{0.9\linewidth}
    \centering
    \centerline{\includegraphics[width=9cm]{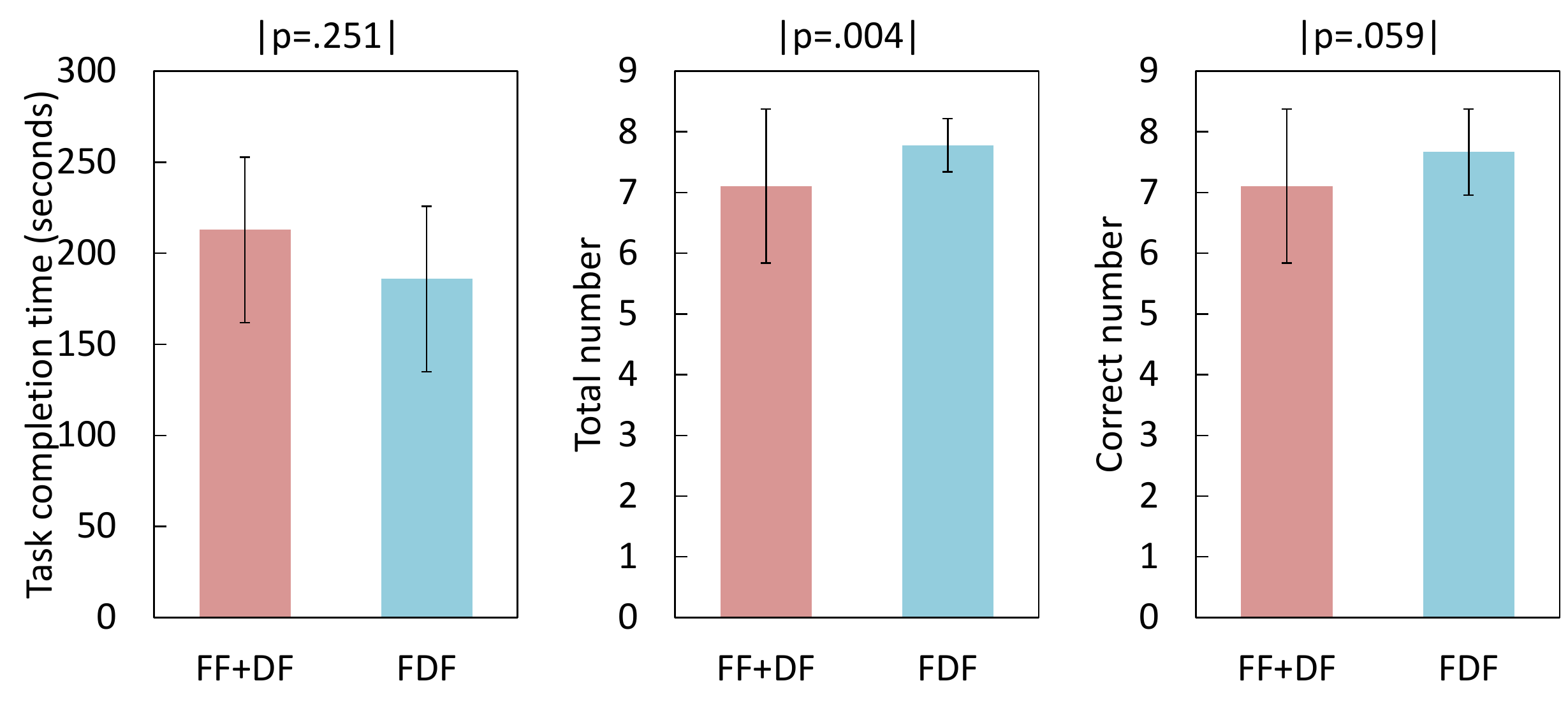}}
    \end{minipage}
}
\subfigure[Subjective measurements]{
    \begin{minipage}[t]{0.9\linewidth}
    \centering
    \centerline{\includegraphics[width=9.1cm]{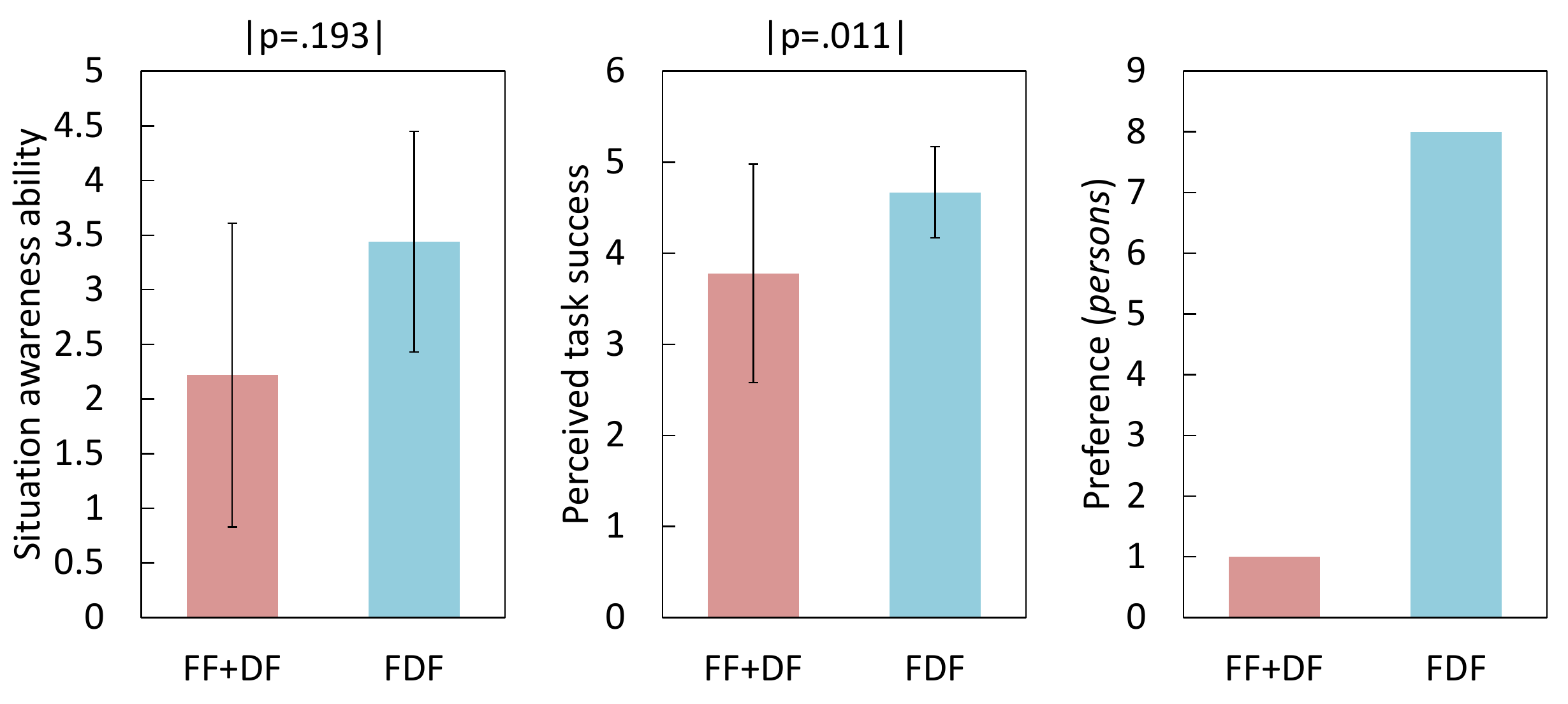}}
    \end{minipage}
}
\caption{Evaluations on the TVUI.}
\label{fig: TIUI_measures}
\end{figure}

\begin{figure}[t]
\centering
\subfigure[Objective measurements]{
    \begin{minipage}[t]{0.9\linewidth}
    \centering
    \centerline{\includegraphics[width=9cm]{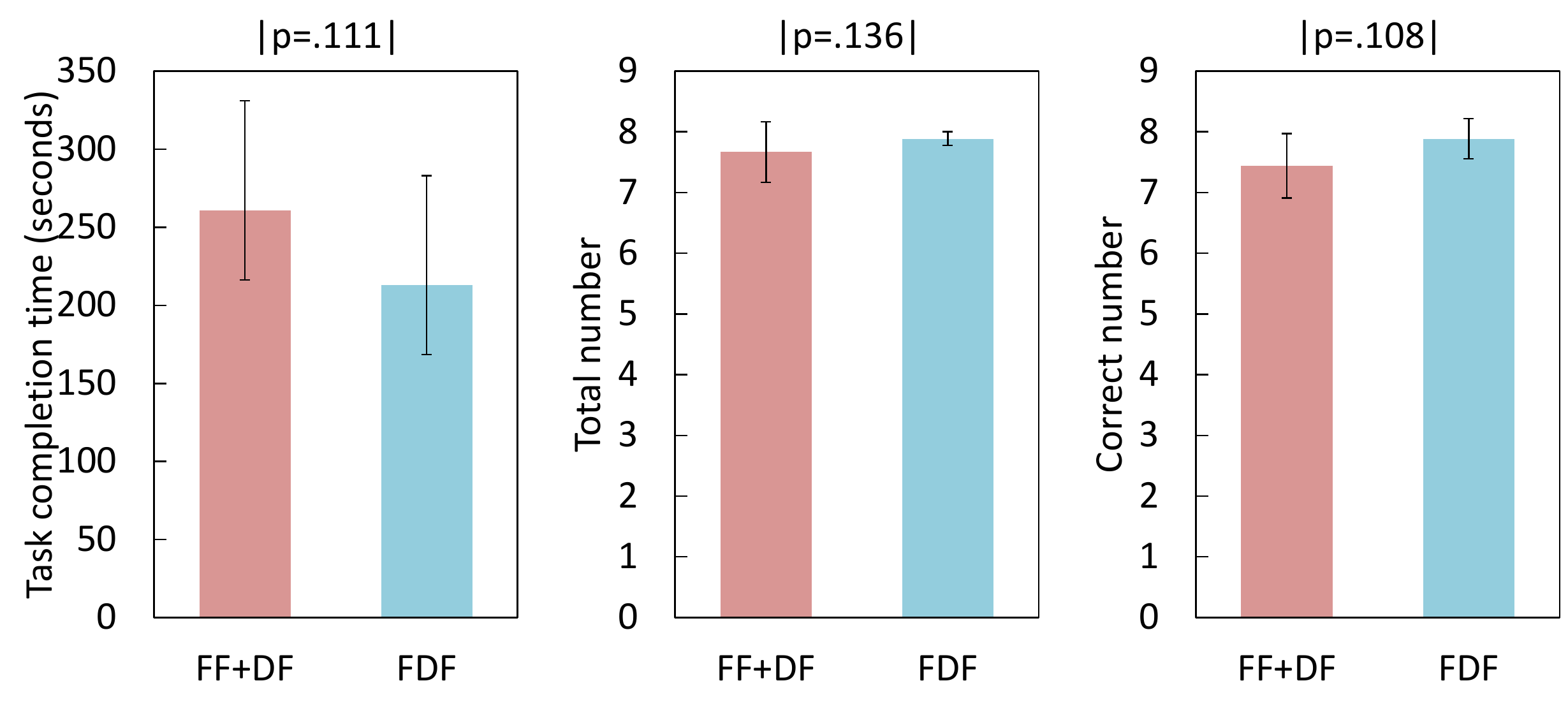}}
    \end{minipage}
}
\subfigure[Subjective measurements]{
    \begin{minipage}[t]{0.9\linewidth}
    \centering
    \centerline{\includegraphics[width=9.1cm]{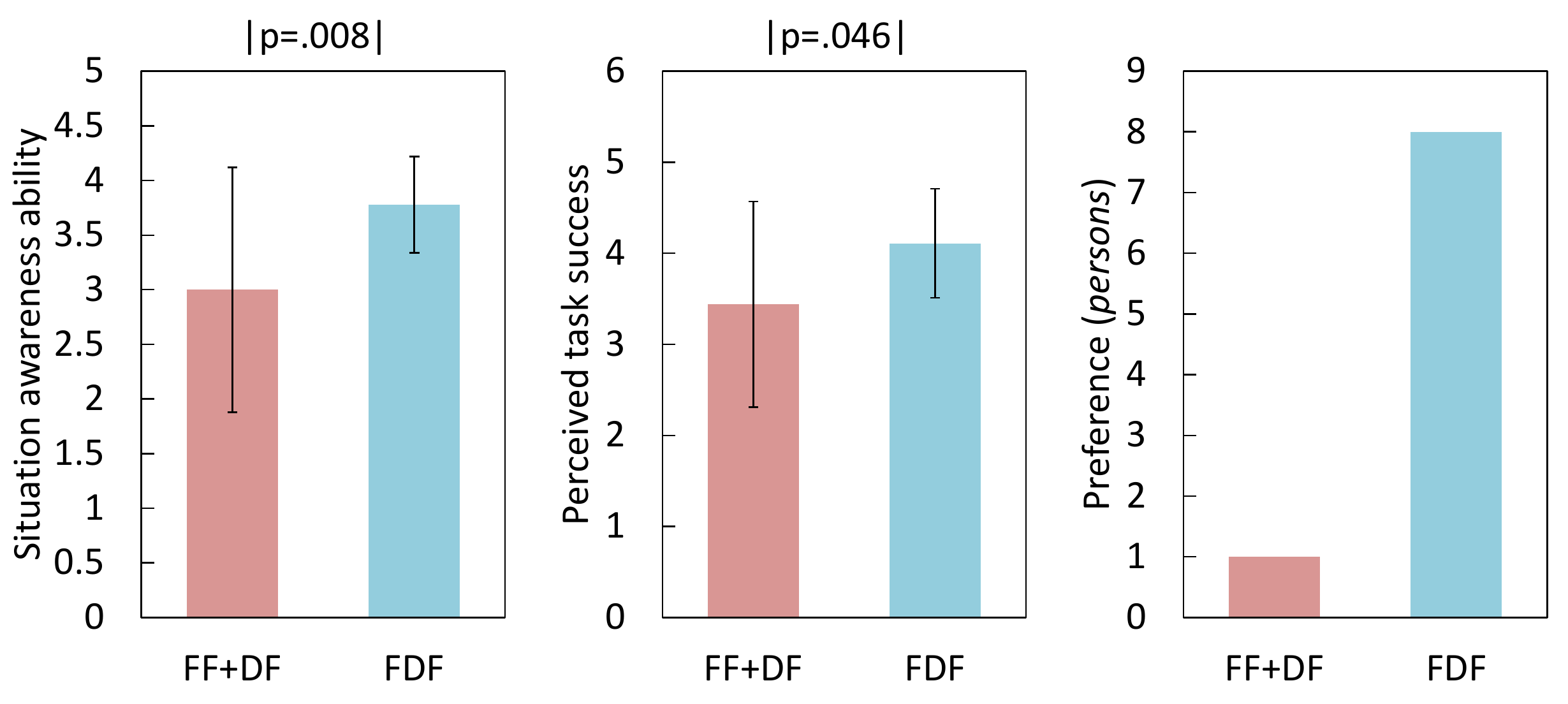}}
    \end{minipage}
}
\caption{Evaluations on the GUI.}
\label{fig: GUI_measures}
\end{figure}
Fig. \ref{fig: TIUI_measures} and Fig. \ref{fig: GUI_measures} show the results of the user study using the TVUI and the GUI, respectively. For simplicity, we denote the user interface with the FDF-video as `FDF', while the user interface with the FF-video and the DF-video as `FF+DF'. The objective measurements include mean task completion time (Task-completion time), the total number of pictures that had been seen (Total number), and the number of correctly recognized pictures (Correct number). The subjective measurements are composed of situation awareness ability, perceived task success, and the participants' preference between the FDF and the FF+DF.

The average task completion time using the TVUI with the FDF was 186.78 seconds (SD=51.082), and 213.44 minutes (SD=39.928) for the FF+DF, as shown in Fig. \ref{fig: TIUI_measures}. The effect on the task completion time was not significant, F(1,16)=0.611, p=0.251, possibly because participants were not skilled in changing the moving speed of the robot. For the pictures that had been seen, the average number was 7.78 (SD=0.441) and 7.11 (SD=1.269) for the FDF and the FF+DF, respectively. We found a significant effect of video stitching on this measurement, F(1,16)=8.286, p=0.004. Besides, the average number of correctly recognized pictures using the FDF (M=7.67, SD=0.707) was larger than using the FF+DF (M=7.11, SD=1.269), and F(1,16)=3.222, p=0.059. The situation awareness ability with the FDF (M=3.44, SD=1.014) was reported to be better than the FF+DF (M=2.22, SD=1.394). There was a significant effect on perceived task success with F(1,16)=5.778, p=0.011, (M=4.67, SD=0.5) for the FDF and (M=3.78, SD=1.202) for the FF+DF. Additionally, almost all participants preferred using the FDF than the FF+DF.

For the GUI, the average task completion time of using the FDF (M=212.89, SD=44.563) was similar with the FF+DF (M=260.67, SD=70.109), and F(1,16)=2.475, p=0.111. The average number of pictures that had been seen using the FDF (M=7.89, SD=0.333) and the FF+DF (M=7.67, SD=0.5) were almost the same. A similar result was achieved on the average number of correctly recognized pictures, (M=7.89, SD=0.333) for the FDF and (M=7.44, SD=0.527) for the FF+DF. The situation awareness ability with the FDF (M=3.778, SD=0.441) was better than the FF+DF (M=3, SD=1.118), F(1,16)=6.429, p=0.008. We also found a significant effect on perceived task success, F(1,16)=3.538, p=0.046, and the FDF had a mean value of 4.11 (SD=0.600) while the FF+DF was 3.44 (SD=1.130). Similar to the user study using the TVUI, almost all participants preferred to using the GUI with the FDF rather than the FF+DF.

Several participants reported that the FDF-video can provide a more compact and high efficient visual feedback than the FF+DF. Using the GUI, six participants said that the user interface with the FDF-video was more convenient in spatial awareness, and they felt easier in distance perception because the scene is continuous. Five participants using the TVUI also gave us the same opinion. Additionally, five participants using the GUI with two separate videos described that they had to frequently switch the two videos, and three operators using the TVUI with two videos noticed the same problem. This problem was not existing on the FDF-video, since the scenes of the upper and the lower part of the FDF-video were continuous.

In summary, both the user study results on the TVUI and the GUI have demonstrated the effectiveness of our method. The telepresence robot incorporating with our video stitching algorithm is able to provide more friendly interactive experiences.
Herein, our work focused on visual feedback with two manners (the stitched video and the two separate videos), instead of the comparison of user interface types (the TVUI and the GUI). The comparison of the TVUI and the GUI can be found in \cite{jia2015telepresence}.

\section{Conclusions}
\label{sec: conclusions}
This paper has proposed to stitch two live videos to provide more compact and high efficient visual feedback for the users of telepresence robots with friendly interactive experiences. The two live videos can be captured by a forward-facing camera with wide-angle lens and a downward-facing camera with fisheye lens for video communication and navigation in robotic telepresence systems, respectively. A multi-homography-based video stitching algorithm, consisting of video image alignment, seam cutting, and image blending, can stitch these videos without calibration, distortion correction, and unwarping procedures. The user study on a telepresence robot was conducted and results demonstrated the effectiveness of our method and the superiority of the telepresence robots with a stitched video as visual feedback.

%\begin{acknowledgements}
%
%\end{acknowledgements}

\section*{Compliance with Ethical Standards}
\label{sec: coi}

\textbf{Funding:} This work was supported by the Natural Science Foundation of China (NSFC) under Grants No. 61773062 and No. 61702037. \\
\noindent\textbf{Conflict of Interest:} The authors declare that they have no conflict of interest.

\bibliographystyle{spbasic}      % basic style, author-year citations
\bibliography{refs}   % name your BibTeX data base

%\begin{thebibliography}{}
%%
%% and use \bibitem to create references. Consult the Instructions
%% for authors for reference list style.
%\end{thebibliography}

\end{document}